\def\year{2020}\relax
\documentclass[letterpaper]{article} 
\usepackage{aaai20}  
\usepackage{times}
\usepackage[square,sort,comma,numbers]{natbib}

\usepackage{lineno}
\usepackage[parfill]{parskip}
\usepackage{multirow}
\usepackage{booktabs}
\usepackage{amsmath} 
\usepackage{helvet} 
\usepackage{courier}  
\usepackage[hyphens]{url}  
\usepackage{amsmath,amssymb}
\DeclareMathOperator{\E}{\mathbb{E}}
\usepackage{graphicx} 
\usepackage{graphicx}  
\title{Continuous Representation of Molecules Using Graph Variational Autoencoder}

\author{\\ \Large \textbf{Mohammadamin Tavakoli, Pierre Baldi}\\\ 
Department of Computer Science,\\ University of California, Irvine\\ 
\{mohamadt, pfbaldi\}@ics.uci.edu 
}
 
\begin{document}

\maketitle

\begin{abstract}
In order to continuously represent molecules, we propose a generative model in the form of a VAE which is operating on the 2D-graph structure of molecules. A side predictor is employed to prune the latent space and help the decoder in generating meaningful adjacency tensor of molecules. Other than the potential applicability in drug design and property prediction, we show the superior performance of this technique in comparison to other similar methods based on the SMILES representation of the molecules with RNN based encoder and decoder.
\end{abstract}

\section{Introduction}
Using machine learning to predict molecular structure properties is a challenging problem \citep{fooshee2018deep, coley2019graph}. While the governing equations (e.g. Schrodinger equation) are difficult and computationally expensive to solve, the fact that an underlying model exists is appealing for machine learning techniques. However, this problem is difficult from a technical point of view. The space of molecules is discrete and non-numerical. Thus, \textit{``how to best represent molecules and atoms for machine learning problems?”} is still a question. 

Despite having numerous ways to represent molecules such as methods introduced in \cite{wu2018moleculenet, bjerrum2017smiles}, all the representations are suffering from a few shortcomings, such as 1) discrete representation, 2) lengthy representation, 3) non-injective mapping, and 4) non-machine readable representation.

Here, we proposed a new method that borrows the main idea from \cite{NIPS2015_5954} and \cite{kipf2016semisupervised} and overcomes all the aforementioned shortcomings. Our method which takes the graphical structure of the molecule as the inputs consists of a variational framework with a side predictor to better prune the structure of the latent space. Then an inner product decoder transfers the samples of latent space into meaningful adjacency tensors. To compare with the main benchmark which is a text-based encoding of molecules \citep{gomez2018automatic} we performed two experiments on the QM9 dataset \citep{ruddigkeit2012enumeration, ramakrishnanquantum} and ZINC \citep{irwin2012zinc}. Both experiments show the success of this method. Although this work is presenting preliminary results of Graph VAE, further experiments and comparisons are left to future work.

\section{Method}
\begin{figure}
  \centering
  \includegraphics[width=\linewidth]{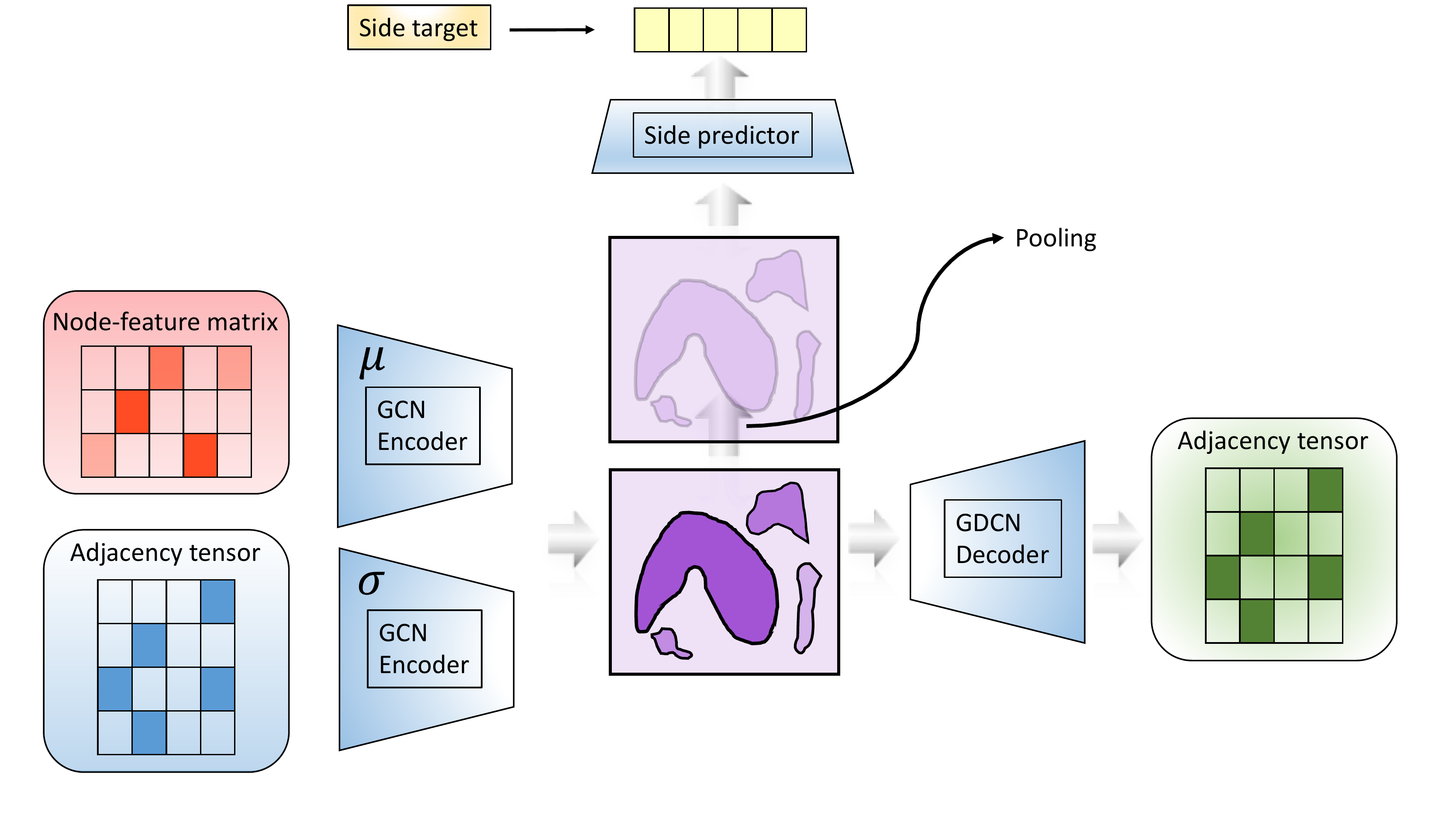}
  \caption{Outline of the model. As depicted above, the model inputs are both node-feature and adjacency tensor, while the model is only outputting the adjacency tensor. The side prediction network is simply using the data points in the latent space as its input.}
  \label{fig:model}
\end{figure}
\paragraph{Molecules and Graphs}
A molecule can be represented by an undirected graph $G = (V, E, R)$, with nodes (atoms) $v_i \in V$ and labeled edges (bonds) $(v_i, e, v_j) \in E$ where $r\in R$ is an edge type. Since we focus on small molecules with four bond types, $R$ is equal to 4. An $(nxd)$ node-feature matrix $H$ is also carrying more information about each node.
These two tensors, together, represent a molecular structure.
\paragraph{Variational Autoencodes}
To help ensure that points in the latent space correspond to valid realistic molecules, and to minimize the dead areas of the latent space, we chose to use a variational autoencoder (VAE). To further ensure that the outputs of the decoder are corresponding valid molecules we employed the open-source cheminformatics suite RDKit30 to validate the chemical structures of output molecules in terms of atomic valence. All invalid outputs are discarded. It is necessary to mention that the ordering of the nodes assumed to be unchanged.
\paragraph{VAE and Side Prediction}

To better learn the graph structure of the molecules, the encoder part of the VAE consists of GCN layers.
The same method as \cite{schlichtkrull2018modeling} has been employed to perform relational update which can be formulated as:

\begin{align*}
    h_i^{l+1} =  \sigma(\sum_{i \in R} \sum_{j \in N_r^{i}} W_r^{(l)}h_j^{(l)} + W_0^{(l)}h_i^{(l)})
\end{align*}

where $N_r^{i}$ denotes the set of nodes connected to node $i$ through the edge type $r \in R$. 
Since we are focusing on small molecules, we applied three layers of GCN in our encoder model to gather information from 3-hop neighbors of each atom.
The structure of encoder consists of two, three-layer GCNs for both mean and the covariance. GCNs of the encoder share the filters of the first two layers.
Here we can formulate the encoding and sampling scheme as follows:
\begin{align*}
    q(\textbf{Z}| H, A) &=  \prod^{N}_{1}q_i(z_i| H, A),\\
    q_i(z_i| H, A) &=  \mathcal{N}(z_i | GCN_{\mu}, GCN_{\sigma})
\end{align*}

The $GCN_{\mu}$ and similarly $GCN_{\sigma}$ are: 
$
    GCN(H, A) = \hat{A}\sigma(\hat{A}\sigma(\hat{A}HW_0)W_1)W_2
$,
where the $\hat{A}$ is the normalized adjacency tensor, $W_i$ is the filter parameter of each layer, and $\sigma$ is the activation function \citep{clevert2015fast}. Finally, as suggested in \cite{kipf2016semisupervised} we use the simplest form of the decoder which can be seen as graph deconvolution network. The output of the encoder is simply the inner product between latent variable:
\begin{align*}
    &p(A|\textbf{Z}) = \prod^{N}_{1}\prod^{N}_{1}p(A_{ij}|z_i, z_j),\\
    &p(A_{ij}=1|z_i, z_j) = \sigma({z_i}^T z_j)
\end{align*}

 For the side prediction part, we employ a simple regression model in the form of a multilayer perceptron (MLP) to the network that predicts the properties from the latent space representation. The input of the side predictor is a vector obtained through a pooling mechanism of the latent representation as follows:
 \begin{align*}
\label{pooling}
G(H^{(L)}) = \sum_{i=1}^{N} softmax(h_i^L.W_p)
\end{align*}
Where $W_P$ is the pooling weight matrix and $H^{(L)}$ is the output of the $GCN_{\mu}$.

Finally, the autoencoder is trained jointly on the reconstruction task and a property prediction task; 
The joint loss function is the summation of the two losses, as follows:
\begin{align*}
    \mathcal{L} &= \textit{ELBO} + \textit{negative log likelihood}\\
    &=\E_{q(\textbf{Z}| H, A)} - KL(q(\textbf{Z}| H, A) || p(Z))\\
    &+ MSE(side network)
\end{align*}

\begin{figure}
  \centering
  \includegraphics[width=\linewidth]{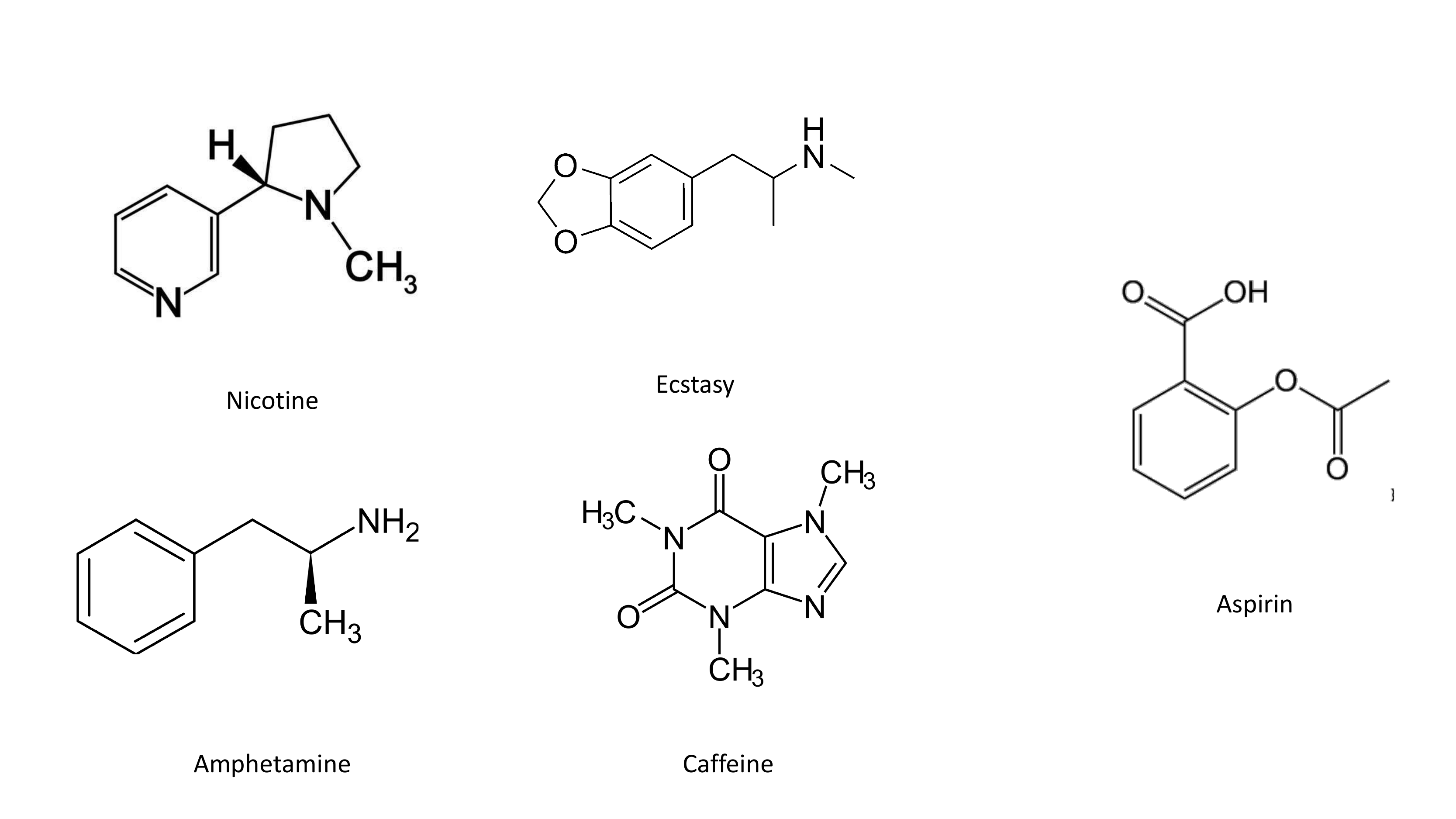}
  \caption{Drugs compare to Aspirin}
  \label{fig:drugs}
\end{figure}

\section{Experiments}
We performed two experiments to show the usefulness of continuous representation. In the first experiment, we focus on the prediction of property and the generation of the valid molecules. In the second experiment, we use this continuous representation to propose a new metric for measuring the molecular similarity.

\subsection{Property Prediction}

\begin{table}[tb]
\scriptsize
\centering
\caption{Three models trained with three different side property. As shown below, using \textit{Druglikeliness} better helps predicting \textit{Solubility} and \textit{Synthesizability}}
 \vspace{2mm}
 \begin{tabular}{ccccc} 
 \toprule
  Side property  & Valid outcome &\textit{Sol}& \textit{Synt} &\textit{Druglikeliness} \\
 \midrule
 Solubility & \textbf{75.3} & 97.03 & 88.7 & 84.2 \\
 Synthesizability & \textbf{73.0} & 89.8 & 98.21& 86.3  \\ 
 Druglikeliness & \textbf{74.6} & 91.0 & 90.7 & 95.11\\
 \bottomrule
 \end{tabular}
 \label{tab:result}
\end{table}
Using a subset of QM9 dataset \citep{ramakrishnanquantum} as the training set, we extract 48,000 molecules covering a broad range of molecules. Each molecule in the training set is chosen to have up to 20 atoms. The training objective on the side predictor was set to be one of the \textit{Solubility}, \textit{Druglikeliness}, and \textit{Synthesizability}. We employ the continuous representation of molecules using each network to predict the other two unseen properties.
The performance of each model plus the percentages of validly generated molecules are summarized in Table \ref{tab:result}.
In order to check the validity of the outcome, we only check for the validity of the atomic valence.
As it is shown in Table \ref{tab:result}
the accuracy of each property is comparable to the state of the art property predictions mentioned in \citep{gilmer2017neural}. Although Graph VAE is not outperforming the predictions based on \cite{gilmer2017neural}, it shows that using a property as a heuristic to prune the latent space, can help with predicting other molecule properties.

\subsection{Molecular Similarity Measure}
Numerous similarity or distance measures have been used widely to calculate the similarity or dissimilarity between two samples. Since metrics are focusing more on 2-dimensional representation rather than 3-dimensional structure, our model as a ``2D structure-aware representation" is an accurate metric for the similarity measure. Normalized Euclidean distance between the latent representation of two molecules after pooling operation is the metric we define to capture the similarity. Here we compare three well-known similarity measures with our technique and also to the methods introduced in \cite{gomez2018automatic}. This method which is using the SMILES representation of the molecules as the input employs a VAE with a side predictor. Both encoder and decoder parts of the VAE are based on RRN and sequence to sequence model. Although all the graphical information of the molecule is encoded within the SMILES representation, inferring the graphical structure (e.g., adjacency tensor) from the SMILES string is an exhausting process that is based on several rules. Despite the numerous techniques built upon using the SMILES representation of the molecules \citep{elton2019deep, hirohara2018convolutional, dalke2018deepsmiles, kwon2017deepcci, krenn2019selfies}, it has been shown that it is more efficient to take advantage of the graph structures and employ GCNs to process molecular structures.  
\begin{table}[tb]

\scriptsize

\centering   
\caption{Similarity measures between Aspirin and four different drugs. Using Graph VAE as a new metrics, shows consistency with other metrics. The GVAE is trained with the solubility as the side property.}
 \vspace{2mm}
 \begin{tabular}{ccccc} 
 \toprule
  metric  & Amphetamine &Ecstasy (MDMA)& Nicotine & Caffeine \\
 \midrule
 Tanimoto & 0.398 & 0.324 & 0.229 & 0.258  \\
 Dice & 0.569 & 0.490 & 0.373 & 0.410  \\ 
 Cosine & 0.607 & 0.490 & 0.374 & 0.434\\
 Graph VAE & 0.363 & 0.199 & 0.147 & 0.176\\
 SMILES VAE \cite{gomez2018automatic} & 0.724& 0.489 & 0.340 & 0.321\\
 \bottomrule
 \end{tabular}
 \label{tab:result2}
\end{table}
Here, we chose Aspirin as a sample drug and compare its similarity with four different drugs with four different similarity measures. 
We compare the performances of our technique with \cite{gomez2018automatic}, which is using a similar approach but operating on text representation of molecules. Our experiment shows that graph-based hidden representation is carrying more information than only text. Table \ref{tab:result2} is summarizing the result of the similarity measure experiment.

As it is shown in table \ref{tab:result2}, our metric is very well aligned with all other well-known metrics which is another proof for the applicability of our model.

\section{Experiment Details}

GVAE consists of two GCNs for the encoder, a pooling mechanism, and a multi-layer perceptron for the side prediction. Both GCNs are three-layer networks with filter matrices $W_0, W_1$, and $W_2$ of 32*32, 32*32m and 32*16 respectively. The pooling weight matrix $W_p$ is of size 1*64 which outputs a vector of length 64 to represent the whole molecule. A two-layer MLP with 32 and 1 hidden units is employed to perform the regression task.\\
\textbf{In Table \ref{tab:result2}}, we use our own implementation of the SMILES VAE. Both GVA and SMILES VAE are trained using a dataset of 70,000 molecules which are randomly selected from ZINC.\\
\textbf{In Table \ref{tab:result2}}, all measures except the continuous representations are calculated with the same fingerprinting algorithm. It identifies and hashes topological paths (e.g. along with bonds) in the molecule and then uses them to set bits in a fingerprint of length 2048. 
The set of parameters used by the algorithm is - minimum path size: 1 bond - maximum path size: 7 bonds - number of bits set per hash: 2 - target on-bit density 0.3.

\section{Conclusion}
We proposed a generative model through which we can find continuous representation for molecules. As shown in the experiments section, this technique can be used in different chemoinformatics tasks such as drug design, drug discovery and property prediction. As future work, one can think of attention based graph convolutions and more complicated decoders. These two extensions can be studied in future works.

\small
\bibliography{template}
\bibliographystyle{abbrv} 

\end{document}